\documentclass{artificial-life}

\usepackage{amsmath}
\usepackage{amssymb}
\usepackage{amsfonts}

\usepackage[style=apa,natbib=true]{biblatex}
\title{Closure of Self-Determining System Based on Causal and Constitutive Relations}

\auth{
    Yoshiyuki Ohmura \affil{1}, 
    Earnest Kota Carr \affil{1},
    Yasuo Kuniyoshi \affil{1} 
}

\corresponding{Yoshiyuki Ohmura}{ohmura@isi.imi.i.u-tokyo.ac.jp}

\affiliations{
    \item The University of Tokyo, 7-3-1 Hongo Bunkyo-ku, Tokyo, Japan
}

\abs{
A self-determining system is defined as one in which causes originating within the system influence the system itself.
This definition raises the question of how to specify 
system boundaries. 
Although the concept of ``closure'' is commonly used for this purpose, defining boundaries solely in terms of causal relations introduce challenges, such as how to handle external causes and circular causality. 
To address this issue, we introduce two types of asymmetric relations: causal and constitutive. We propose that system boundaries can be defined as closures of loops formed by these relations, referred to as causal-constitutive loops. By constraining constitutive relations, the resulting system necessarily includes internal causes and thereby satisfies self-determination. Furthermore, to prevent reduction to supervenience, constitutive relations must involve at least two independent variables. This minimal requirement leads to two interdependent loops, which implies a dual-process organization.
}


\begin{document}

\coverpage           %
\doublespacing       %


\section{Introduction}
Autonomy is a key concept in artificial life and is often associated with self-determination (\cite{boden}), which refers to the idea that the causes of a system’s actions originate within the system itself (\cite{haggard2018, ohmura2026}). Although autonomy has been examined in several contexts, including metabolism, behavior, and cognition (\cite{barandiaran2008, kolchinsky2018}), this study treats it as a broad organizing concept and focuses specifically on self-determination as one of its core aspects. 

A central challenge in this context is how to define the boundary of such a system. In biological and artificial life research, this problem has frequently been approached using the concept of closure (\cite{bowden2020}). 
However, many of these approaches focus primarily on the self-maintenance of biological systems rather than on the origin of internally generated causes.

Several variants of closure have been proposed, including closure of efficient causes (\cite{rosen, hofmeyr}), 
closure of cognitive operational structure (\cite{piaget1971biology}), 
operational closure (\cite{maturana}), 
closure of self-generative transformation network (\cite{varela}), 
closure of constraints (\cite{montevil2015, mossio2017}), 
semantic closure (\cite{pattee,amahury}), 
and catalytic closure (\cite{kauffman}). 
However, these approaches typically characterize closure using only one type of asymmetric relation, which makes it difficult to capture key requirements of self-determining systems, in particular the simultaneous presence of internally originating causes and openness to external influences.
This is because self-determination inherently requires distinguishing internal from external causes.

In contrast, we argue that defining system boundaries in the context of self-determination requires at least two distinct types of asymmetric relations, namely causal and constitutive relations. Causal relations describe temporal asymmetry between cause and effect, whereas constitutive relations describe simultaneous dependence between a whole and its parts. Based on this distinction, we propose that system boundaries can be defined as closures of loops formed by these two types of relations.

The framework developed in this study satisfies three requirements:
\begin{enumerate}
\item the system remains open to external causal influences. 
\item its boundary is defined through well-defined closure.
\item circular causality alone is avoided. 
\end{enumerate}
To achieve this, we introduce causal–constitutive loops (CC-loops), in which causal processes 
and constitutive dependencies jointly form closed structures.

By placing constraints on constitutive relations, we demonstrate that such systems necessarily include internal causal relations, thereby meeting the condition of self-determination. Furthermore, to avoid reduction to supervenience, constitutive relations must involve at least two causally independent variables, which results in multiple CC-loops as the minimal structural requirement.

Our approach is related to the tradition of autopoiesis and organizational closure (\cite{maturana}). However, rather than addressing self-maintaining or living systems, we focus on self-determination as the presence of internally originating causes of action and propose a minimal formal structure for such systems.

\section{Causality and Circularity}
Causality is an asymmetric relationship between cause and effect and is defined as follows.
\begin{equation}
b:=F(a).
\end{equation}
where $a$ denotes the cause, $b$ is the effect, and $F$ denotes the causal transmission mechanism. $:=$ is the causal assignment operator (\cite{pearl2009}). 
According to the interventionist account of causation (\cite{pearl2009, woodward2003}), causal asymmetry exists when a change in $a$ produces a change in $b$, while a change in $b$ does not alter $a$. 
Similarly, an asymmetry between $F$ and $b$ implies that can also be subject to intervention $F$. 
We distinguish $a$ as a ``variable-level cause'' and $F$ as a ``structure-level cause.'' In this study, lowercase letters denote multidimensional vector variables, whereas uppercase letters denote functions.

If closure is defined only in terms of causal relationships, it necessarily entails circular causality. However, circular causality is conceptually problematic. In cases of steady-state circularity, it produces a ``chicken-and-egg'' dilemma and does not allow a clear identification of an initiating cause of action. Although transient circular causality cannot be excluded, it is too unstable to serve as a basis for defining system boundaries. Consequently, it is natural to introduce additional asymmetric relationships beyond causal ones.

\section{Constitution and Supervenience}
In this study, we define the ``constitution'' relation as follows.
\begin{equation}
    F:\sim \mathcal{G}(c_j), j \in J \subset \mathbb{N}.
\end{equation}

This relationship implies that $F$ is composed of a finite number of $c_j$.
A change in $F$ requires a corresponding change in $c_j$. 
Conversely, changes in $c_j$ does not necessarily lead to changes in $F$, reflecting multiple realizability.
We argue that, since $c_j$ and $F$ share the physical entities, they can serve as criteria for determining the identity of a system.

We allow that causal transmission mechanisms (functions) may be constituted by variables, but we do not assume that variables are constituted by other variables. If a variable were taken to be constituted by other variables, the relation would reduce to a definitional equivalence of the form $a=(b,c)$, without introducing any additional dependencies. Therefore, treating variables as constituted in this way amounts to a theoretically trivial reformulation, and we do not adopt it as a concept of constitution in this study.

Our concept of constitution differs from that of production. It is a mereological relation similar to the way a clock is constituted by gears, motors, and hands. Changes in the parts occur simultaneously with changes in the whole. This relation is not temporal in nature; rather, when the whole changes, its constituent parts change simultaneously. Such changes in causal transmission mechanisms correspond to structure-level causes.

When a constitutive relation is specified using a single variable, it effectively collapses into supervenience. In such cases, a supervenient entity cannot change without a corresponding change in the subvenient entities (\cite{kim1998}).
Considering a composition involving a single variable (vector), if $F: \sim \mathcal{G}(c)$ holds, 
then any change in $F$ necessarily entails a change in $c$. 
Therefore, these special cases of constitutive relations are subsumed under supervenience relations. We argue that it is important to explicitly distinguish between supervenience and constitutive relations. This is because, as shown below, it is well established that constructing a cycle that combines supervenience and causal relations results in the exclusion problem (\cite{kim1998}).

\section{Supervenience and Exclusion}
A well-known tension exists between supervenience and causality, a topic extensively discussed in the philosophy of mind (\cite{kim1998}). The exclusion argument holds that a supervenient entity cannot exert causal influence on the subvenient entities on which it depends.

When $F$ supervenes on $c$, as in $F:\sim \mathcal{G}(c)$,
a causal relation between $F$ and $c$ should not be postulated. 
In this case, any change in $F$ necessarily depends on a change in $c$. 
However, causal asymmetry requires that changes in $F$ 
do not bring about changes in $c$. 
Therefore, these two requirements are incompatible.

This implies that loops cannot be constructed using only supervenience and causal relations. In other words, a constitutive relation must involve at least two variables that are independently manipulable. We adopt the requirement that all variables constituting the causal transmission mechanism participate in loops formed through causal and constitutive relations as a criterion for defining the boundary of a self-determining system.

\section{Simple Self-determining System}
A CC-loop is a causal relationship in which changes in the causal transmission mechanism $F$ serve as the structure-level cause, and the outcome affects the variables that constitute $F$, thereby forming a closed loop.

\begin{figure}[htb]
\begin{center}
\includegraphics[width=8cm]{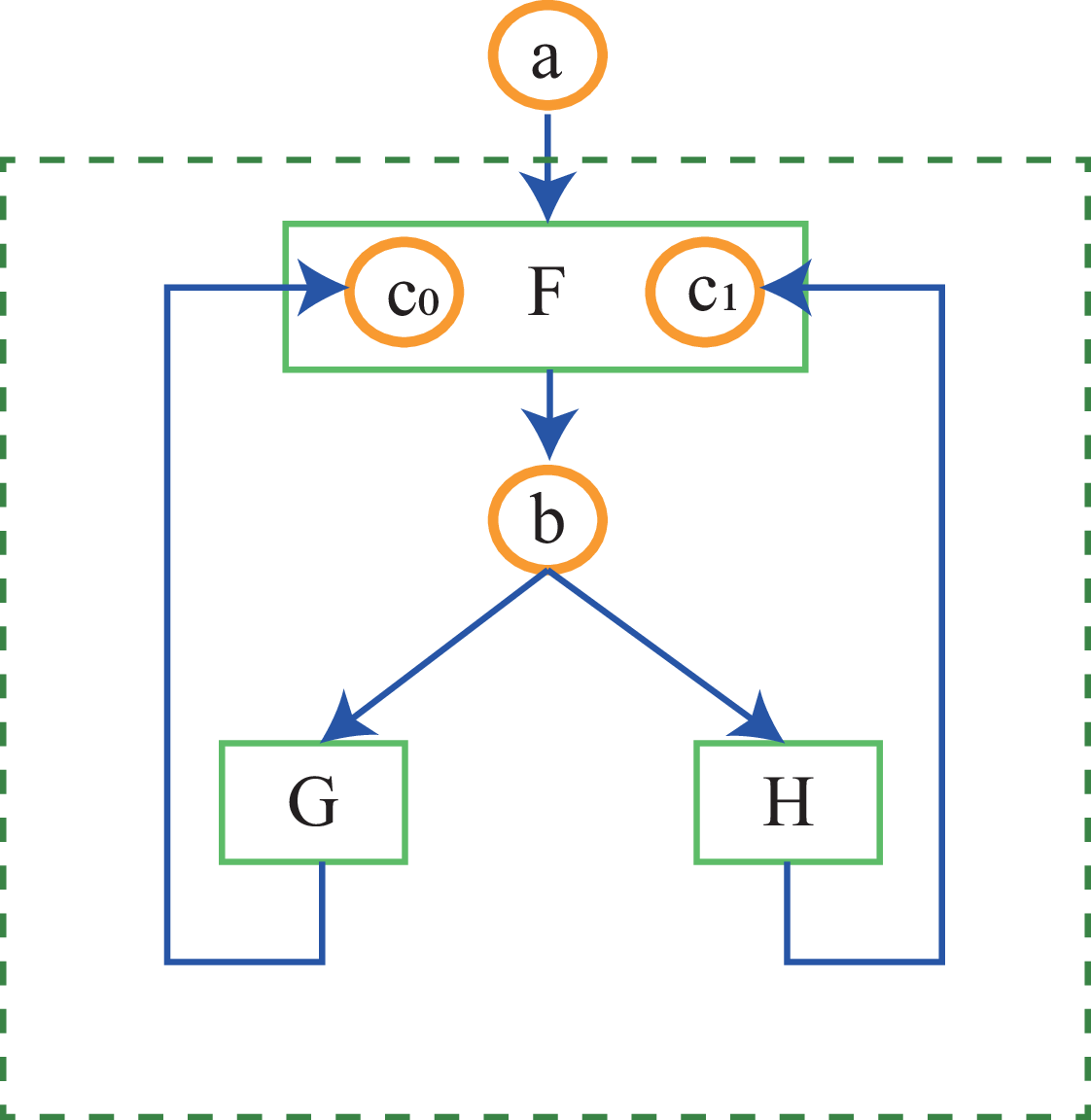}
\caption{Minimal self-determining system defined by a causal–constitutive loop (CC-loop). 
Blue arrows indicate causal relations. 
Green boxes represent causal transmission mechanisms (functions), and orange circles 
represent variables. 
The orange circle inside the green box indicates a compositional relation.
The mechanism $F$ generates an effect $b$ from an input $a$    ($b := F(a)$). 
The variable $b$ exerts causal causal influence on two internal variables $c_0$ and $c_1$, which jointly constitute 
the mechanism $F$. 
This structure yields two closed CC-loops:
 $F \to b \to c_0 \leadsto F$ and $F \to b \to c_1 \leadsto F$.
These loops ensure that changes in $F$ propagate through $b$ and subsequently return to modify the structure of
$F$ via its constitutive variables.
Consequently, the system contains internal causes that act on its own causal structure, thereby satisfying self-determination. The input $a$ functions as as an external cause, since it is not part of any CC-loop. Therefore, the system remains open to external influences while maintaining an internally closed internal organization.
The requirement that $F$ is constituted by at least two independent variables ($c_0$, $c_1$) prevents reduction to a supervenience relation and ensures the existence of multiple independent CC-loops. Thus, causal transmission ultimately affects the variables that constitute $F$.
}
\label{fig:simple_model}
\end{center}
\end{figure}

The following example describes the simplest self-determination system (see also Fig. \ref{fig:simple_model}). 

\begin{gather}
    b:=F(a) \label{cause1} \\
    c_0 := G(b) \label{cause2} \\
    c_1 := H(b)  \label{cause3} \\
    F :\sim \mathcal{B}(c_0, c_1) \label{const}
\end{gather}

Equations (\ref{cause1})--(\ref{cause3}) specify
causal relations. 
The mechanism $F$ is constituted by two variables: $c_0$  and $c_1$.
If $F$ were defined by only a single variable, it would 
reduce to a supervenience relation. 
In that case, it  would be impossible to construct a loop by combining causal and supervenience relations. Therefore, two CC-loops can be identified: $F \to b \to G \to c_0 \leadsto F$ and $F \to b \to G \to c_1 \leadsto F$ where $\leadsto$ represents constitutive relation and $\to$ denotes causal relation. 

\section{External Causes}
External causes are defined as influences that originate outside a system. A self-determining system cannot be defined simply by its isolation from external causes, since many organisms are affected by sensory information from the outside. Therefore, specifying the boundary of a system necessarily requires identifying and delimiting influences that arise from outside the system.

In the system described above, $a$ functions as an external cause. 
A key feature of an external cause is that its causal or constitutive relations are not connected to variables or functions within the system. In Fig. \ref{fig:simple_model}, $a$ does not participate in any CC-loop.

\section{Boundary of Self-determining System}
In this study, we consider a set $U$ of objects comprising both variables and functions. 
In this context, the system interior $I$ is defined as a subset of $U$ ($I \subset U$). 
Influences originating outside the system $\overline{I} $ are referred to as external causes.

For any given $x \in I$, its existence is justified by the elements in $I$ as follows:
\begin{equation}
b \in I \Rightarrow \exists F \in I, \exists a: b:=F(a) \label{cond_1} \\
\end{equation}

\begin{equation}
F \in I \Rightarrow 
\left\{ 
\begin{split}
\exists a, b \in I: b:=F(a) \\
F :\sim \mathcal{B}(c_j), \forall j \in J: c_i \in I 
\end{split}
\label{cond_2}
\right.
\end{equation}    

Formula \ref{cond_1} specifies 
the condition that must hold for a variable to be considered internal to the system.
We do not require that $a$ be internal to the system, since  external causes are permitted. 
In this setting, if $F$ is internal to the system, then changes in $F$ can function as a structure-level cause, which implies that  $b$ is included within the system.

Formula \ref{cond_2} specifies the conditions that must be met when the causal transmission mechanism belongs to the system.
For a causal transmission mechanism to be included in a system, it must satisfy one of two criteria.
The first criterion is that if both the cause $a$ and the effect $b$ are variables within the system, then $F$ is also considered to be part of the system. In this case, external causes other than $b$ are allowed in this formula.
The second criterion is that all causal variables constituting $F$ must be contained within the system.

Since we exclude loops that only rely on causal relations, we construct loops using constitutive relations. However, when constitution is reduced to supervenience, it becomes incompatible with causal relations; therefore, the number of elements in $J$ must be at least two. In this case,
$c_0$ and $c_1$ are required to belong to the system.
The boundary of a self-determining system is then defined according to these conditions.

A self-determining system contains at least two CC-loops. From condition 1, all variables must satisfy causal requirements to be included within the system, which implies that a self-determining system necessarily includes intrinsic structure-level causes. In addition, to avoid loops formed exclusively by causal relations, the system must incorporate causal and constitutive relations.

\section{Prohibitions}
When forming loops using causal and constitutive relations, several structures must be excluded.
\begin{enumerate}
    \item Constitutive relations are allowed only with respect to causal transmission mechanisms. 
    \item  Loops constructed exclusively from causal relations are not permitted. 
    \item  The constitutive relation must not reduce to supervenience relation.
    \item In a causal relationship $b := F(a) $, the effect variable $b$ does not constitute its causal transmission mechanism $F$.
\end{enumerate}

We have already discussed Prohibitions 1--3. 
Prohibition 4 arises from the fact that causal asymmetry 
cannot be consistently defined even when causal and constitutive relations form a local cycle. 
If a change in $b$ leaves $F$ invariant, this does not contradict causal asymmetry; however, we consider this condition to be impractical, as it renders the role of $b$ within the local causal–constitutive loop unclear. 

\section{Causality among constitutive variables}
In the simple self-determination system described above, when $F :\sim \mathcal{B}(c_0, c_1)$ holds, no causal relationship can be defined between $c_0$ and $c_1$.
This is because if $c_0$ changes as a result of $c_1$, then $F$ will also change via $c_0$ unless $F$ is invariant under variations in $c_0$.
If a causal relationship such as $F:=I(c_0)$ is introduced, a circular causality emerges: $F \to b \to G \to c_0 \to I \to F$. 
This would undermine the constitutive relationship that was introduced to avoid circularity based solely on causal relations.
Consequently, causal relations should be disallowed between variables that jointly constitute the same causal transmission mechanism.

In our definition of a self-determining system, cycles formed exclusively by causal relations are excluded, and constitutive relations are restricted to causal transmission mechanisms. Therefore, the possible structures of self-determining systems are significantly constrained. Conversely, although system boundaries are determined through CC-loops, the system is not isolated, since it explicitly accommodates external causes.

\section{Conclusions}

In this study, we introduced a framework for defining the boundary of a self-determining system based on the closure of causal and constitutive relations. Unlike existing closure frameworks based on a single type of relation, our approach combines these relations into CC-loops, enabling system boundaries that permit external causes while ensuring that internal causes affect the system itself. 

Within the specific context of self-determination as the problem of internally originating causes rather than self-maintenance, this integration avoids difficulties associated with circular causality and incompatibility with supervenience. We showed that constitutive relations must include at least two independent variables, yielding multiple interdependent CC-loops as the minimal structure for self-determination. 

Future studies will examine the implications of this framework for specific biological and cognitive systems, in particular by extending the notion of constitutive relations to cases where causal transmission mechanisms are generated through material interactions within the system.

\printbibliography

\end{document}